\documentclass{article}

\PassOptionsToPackage{table}{xcolor}

 \usepackage[main, final, nonatbib]{neurips_2026} 

\usepackage[para]{footmisc}
\usepackage{makecell}
\usepackage{array}
\newcolumntype{C}[1]{>{\centering\arraybackslash}p{#1}}
\usepackage[utf8]{inputenc} 
\usepackage[T1]{fontenc}    
\usepackage{hyperref}       
\usepackage{url}            
\usepackage{wrapfig}
\usepackage{booktabs}       
\usepackage{multirow}
\usepackage{graphicx}
\usepackage{amsmath}
\usepackage{amsfonts}       
\usepackage{nicefrac}       
\usepackage{microtype}      
\usepackage{xcolor}         

\definecolor{timebg}{RGB}{232,240,250}
\definecolor{oursbg}{RGB}{255,250,210}

\title{Focus Matters: Attention-Value Dynamics for Efficient Hallucination Mitigation in Vision-Language Models}

\author{%
  \textbf{Sohyeon Kim}$^{1}$
  \quad
  \textbf{Sang Yeon Yoon}$^{2}$
  \quad
  \textbf{Kyeongbo Kong}$^{1, 3}$\thanks{Corresponding author.}\\[0.4em]
  $^{1}$ Dept. of Electrical \& Electronics Eng., Pusan National University\\
  $^{2}$ Dept. of Computer Eng. \& AI, Pukyong National University\\
  $^{3}$ AI Collaboration Hub Research Center, Pusan National University\\
  Republic of Korea\\[0.2em]
  \texttt{\{shkim0503,kbkong\}@pusan.ac.kr}
  \quad
  \texttt{palsc@pukyong.ac.kr}
}

\begin{document}

\maketitle

\begin{abstract}
Large Vision-Language Models (LVLMs) have achieved impressive progress in multimodal reasoning, yet they remain prone to object hallucinations, generating descriptions of objects that are not present in the input image. In this work, we investigate hallucination from the perspective of attention-value dynamics inside LVLM vision encoders. We identify a consistent three-phase structure of visual processing---diffusion, focus, and rediffusion---and show that the focus phase is where attention most clearly separates strongly and weakly supported visual tokens. However, low attention does not necessarily imply negligible downstream influence: low-attention tokens in the focus phase can still exert non-negligible value-side influence on the attention output relative to their small attention mass.
Through controlled phase-wise value interventions, we find that hallucination behavior is particularly sensitive to the value content of low-attention tokens during this phase. Replacing or neutralizing these values reduces hallucination metrics while largely preserving grounded object evidence. A token-level teacher-forcing analysis further shows that the intervention reduces the probability of hallucinated object tokens with much smaller effects on ground-truth object tokens. In addition, Visual Attention Ratio (VAR) analysis shows that focus-phase intervention is accompanied by increased attention to visual tokens during decoding. Based on these observations, we instantiate a simple training-free inference-time intervention that replaces focus-phase low-attention values with an image-level mean value vector using statistics from a single forward pass. Experiments across multiple LVLM backbones demonstrate that this analysis-derived intervention reduces object hallucination with negligible additional runtime and remains compatible with existing inference-time mitigation methods. Our project page is available at: \url{https://paper.pnu-cvsp.com/FocusMatters/}
\end{abstract}

\section{Introduction}

Large Vision-Language Models (LVLMs) have achieved remarkable progress in multimodal understanding and image-grounded text generation~\cite{llava,llava1_5,shikra,internVL2_5,Qwen2.5-VL}.
Despite these advances, they remain vulnerable to \emph{object hallucination}~\cite{CHAIR,survey_hallu}, generating descriptions of objects that are not present in the input image.
Such hallucinations undermine the reliability of LVLMs and limit their use in applications requiring faithful visual grounding. This issue becomes increasingly important as LVLMs are incorporated into multimodal agents. In such settings, visual outputs may influence subsequent reasoning, planning, tool use, or inter-agent communication, and unreliable visual information can propagate across multiple stages of an agentic pipeline, amplifying the impact of an initial hallucination. At the same time, agentic systems often invoke multimodal models repeatedly across multiple reasoning steps or agents, making computational efficiency increasingly important. Therefore, as LVLMs are deployed in more complex multimodal and agentic systems, both reliable visual grounding and efficient inference become essential.

Existing hallucination mitigation methods largely intervene during decoding or within the language model.
Decoder-stage methods adjust token probabilities, contrast visual conditions, or modify generation dynamics~\cite{DoLA,VCD,Opera}, while language-side attention methods reweight the model's reliance on visual tokens during generation~\cite{PAI,Devils}.
Although effective, these methods operate after visual features have already been extracted, without directly addressing how individual visual tokens contribute to hallucination. Recent studies on visual token pruning~\cite{baek2026agilepruner} have also shown that selectively removing certain visual tokens can mitigate hallucination, suggesting a potential connection between unreliable visual information and hallucination behavior.
Vision-encoder-side methods intervene earlier by suppressing unreliable visual signals before they reach the language model.
A representative approach, \textit{Adversarial Uncertainty Estimation} (AUE)~\cite{AUE}, identifies uncertain visual tokens through iterative adversarial optimization using Projected Gradient Descent (PGD)~\cite{PGD}.
However, this per-image optimization introduces substantial latency; on Qwen-2.5-VL, for example, AUE increases total inference time from 2.60s to 33.46s, making it approximately 12.9$\times$ slower than standard inference.
This motivates a closer analysis of where hallucination-sensitive visual signals arise inside the vision encoder and whether they can be controlled without iterative optimization.

In this work, we study object hallucination through the lens of \emph{attention-value dynamics} inside LVLM vision encoders.
Self-attention propagates visual information by aggregating value vectors according to attention weights, so attention scores alone do not fully characterize a token's downstream influence.
As summarized in Fig.~\ref{fig:attn_value_dynamics}, we observe a consistent three-phase structure across vision encoder layers: \emph{diffusion}, \emph{focus}, and \emph{rediffusion}.
Attention is broadly distributed in the diffusion phase, becomes concentrated on a small subset of tokens in the focus phase, and becomes diffuse again in the rediffusion phase.

The focus phase is important because it is where attention most clearly separates strongly supported tokens from weakly supported ones.
However, low attention does not necessarily mean that a token's value content is removed from the computation.
We find that focus-phase low-attention tokens receive small attention mass but still exert non-negligible value-side influence on the attention output.
These tokens do not dominate the visual representation, but their value contribution is disproportionately large relative to their attention mass.
This attention-value mismatch raises the question of whether weakly supported value-side signals are related to hallucinated object generation.

To examine this question, we conduct controlled phase-wise value interventions.
We use value modification as a diagnostic tool to test how different value groups affect hallucination behavior.
Modifying low-attention values during the focus phase reduces hallucination metrics while largely preserving caption fidelity, whereas analogous interventions in the diffusion or rediffusion phases have weaker effects.
In contrast, suppressing high-attention values substantially harms caption fidelity, indicating that strongly attended values contain important grounded visual evidence.
This contrast suggests that hallucination reduction is not simply caused by removing visual information, but is closely related to neutralizing weakly supported value-side signals in the focus phase.

We further validate this observation using token-level teacher forcing.
For hallucinated and ground-truth object tokens identified from generated captions, we recompute token log-probabilities under the same token sequence while applying focus-phase value interventions.
The intervention reduces hallucinated object token probabilities while having much smaller effects on ground-truth object probabilities.
Visual Attention Ratio (VAR) analysis~\cite{Devils} further shows that this intervention is accompanied by increased attention to visual tokens during decoding.
Together, these results provide diagnostic evidence that focus-phase low-attention value signals are associated with hallucination-sensitive visual information.

Based on these findings, we instantiate a simple analysis-derived intervention: focus-phase low-attention mean value replacement.
The method identifies low-attention tokens using attention statistics from a single forward pass and replaces their value vectors with the image-level mean value vector.
This neutralizes token-specific value deviations without removing tokens, requiring additional training, or performing adversarial optimization.
Experiments across multiple LVLM backbones show that this intervention reduces object hallucination with negligible runtime overhead and remains compatible with existing inference-time mitigation methods.

Our contributions are summarized as follows:
\begin{itemize}
    \item \textbf{Attention-value dynamics analysis.}
    We identify a diffusion--focus--rediffusion structure in LVLM vision encoders and show that the focus phase exhibits an attention-value mismatch: low-attention tokens receive small attention mass while retaining non-negligible value-side influence.

    \item \textbf{Diagnostic evidence for hallucination-sensitive value signals.}
    Through phase-wise value interventions and token-level teacher-forcing analysis, we provide diagnostic evidence that focus-phase low-attention value signals affect hallucinated object likelihood under controlled interventions.

    \item \textbf{Analysis-derived lightweight intervention.}
    We instantiate a training-free intervention that replaces focus-phase low-attention values with an image-level mean value vector using statistics from a single forward pass.

    \item \textbf{Empirical validation across LVLM backbones.}
    We evaluate the intervention across multiple LVLMs and show a favorable trade-off between hallucination reduction, caption fidelity, and inference efficiency.
\end{itemize}

\section{Related Work}
\label{sec:related_work}

\paragraph{Large Vision-Language Models.}
Large Vision-Language Models (LVLMs)~\cite{llava,llava1_5,shikra,internVL2_5,Qwen2.5-VL} extend large language models (LLMs)~\cite{touvron2023llama,touvron2023llama2,chiang2023vicuna} with visual inputs.
A typical LVLM consists of a vision encoder~\cite{dosovitskiy2020vit,CLIP,zhai2023sigmoid}, a modality connector, and an LLM.
Despite strong multimodal performance, LVLMs remain prone to object hallucination, generating objects or relationships that are inconsistent with the input image.

\paragraph{Hallucination Mitigation in LVLMs.}
Hallucination mitigation methods can be broadly divided into training-based and training-free approaches.
Training-based methods use hallucination-aware supervision or preference optimization~\cite{MMhal-Bench,zhao2023beyond,xie2024v,yang2025mitigating,HDPO}, but require additional data and training.
Training-free methods instead intervene during inference.
Decoder-side methods modify token probabilities or generation dynamics~\cite{DoLA,VCD,Opera}, while language-side attention methods adjust the decoder's reliance on visual tokens~\cite{PAI,Devils}.
These methods are complementary to ours but operate after visual features have already been extracted.
Vision-encoder-side methods such as AUE~\cite{AUE} intervene earlier by estimating unreliable visual tokens through iterative PGD-based optimization~\cite{PGD}, which introduces substantial latency.
In contrast, our work analyzes attention-value dynamics inside the vision encoder and applies a single-pass, training-free intervention during the focus phase.

\section{Attention-Value Dynamics in Vision Encoders}
\label{sec:attention_value_dynamics}

To identify where hallucination-sensitive visual signals arise inside LVLM vision encoders, we analyze how visual information is routed and transformed across layers. In self-attention, attention weights determine which tokens are referenced, while value vectors determine what information is propagated. Therefore, attention scores alone do not fully characterize a token's downstream influence. In particular, a token receiving low attention may still affect the attention output if its value content contributes non-negligibly to the weighted aggregation. Motivated by this distinction, we jointly analyze layer-wise attention concentration and low-attention value influence across vision encoder layers.

\subsection{Three-Phase Structure from Attention Concentration}
\label{sec:phase_structure}

\begin{figure}[t]
\centering
\hspace*{0cm}
\includegraphics[width=\textwidth]{att_dynamics_main.pdf}

\caption{
\textbf{Attention-value dynamics across vision encoder layers.}
(a) Layer-wise attention concentration ratio $R^{(\ell)}$ reveals a consistent diffusion--focus--rediffusion structure.
(b) Low-attention value influence across layers.
We measure the attention mass $\alpha_{\mathcal{L}}$ assigned to low-attention tokens, their normalized value contribution $\eta_{\mathcal{L}}$, and the influence per attention mass $G_{\mathcal{L}}=\eta_{\mathcal{L}}/(\alpha_{\mathcal{L}}+\epsilon)$.
During the focus phase, $\alpha_{\mathcal{L}}$ decreases sharply while $\eta_{\mathcal{L}}$ remains non-negligible, yielding a large $G_{\mathcal{L}}$.
This indicates that low-attention tokens can exert disproportionate value-side influence relative to their small attention mass.
}
\vspace{-10pt}
\label{fig:attn_value_dynamics}
\end{figure}

We first quantify how attention distributions evolve across layers.
For a given layer $\ell$ and attention head $h$, let $\mathbf{A}^{(\ell,h)} \in \mathbb{R}^{N \times N}$ denote the self-attention matrix over visual tokens.
For architectures with a CLS token, we define the reference attention distribution from the CLS token to the $N$ spatial tokens as
\begin{equation}
a^{(\ell,h)}=\{A^{(\ell,h)}_{\mathrm{cls},i}\}_{i=1}^{N}.
\end{equation}
For architectures without an explicit CLS token, such as Qwen-2.5-VL, we instead use the average attention distribution over all visual query tokens:
\begin{equation}
a_i^{(\ell,h)}
=
\frac{1}{N}
\sum_{j=1}^{N}
A_{j,i}^{(\ell,h)} .
\end{equation}

To measure attention concentration, we use two statistics: attention entropy and maximum attention score.
The attention entropy is defined as
\begin{equation}
H^{(\ell,h)}
=
-\sum_{i=1}^{N}
a_i^{(\ell,h)}
\log a_i^{(\ell,h)},
\end{equation}
which measures how broadly attention mass is distributed across visual tokens.
The maximum attention score is defined as
\begin{equation}
M^{(\ell,h)}
=
\max_i a_i^{(\ell,h)},
\end{equation}
which captures the dominance of the most attended token.
We summarize layer-wise attention concentration as
\begin{equation}
R^{(\ell)}
=
\frac{\mathbb{E}_{h}[M^{(\ell,h)}]}
{\mathbb{E}_{h}[H^{(\ell,h)}]} .
\end{equation}
A larger $R^{(\ell)}$ indicates that attention is concentrated on a small subset of tokens, whereas a smaller value indicates a more diffuse attention distribution.

As shown in Fig.~\ref{fig:attn_value_dynamics}(a), the layer-wise evolution of $R^{(\ell)}$ reveals a consistent three-phase structure.
Although the absolute magnitude differs across models, the depth-wise pattern remains similar: attention is broadly distributed in early layers, becomes sharply concentrated in intermediate layers, and becomes more diffuse again in later layers.
Based on this pattern, we divide the vision encoder into three phases:

\begin{itemize}
    \item \textbf{Diffusion phase.}
    In the early layers, $R^{(\ell)}$ remains low, indicating that attention is broadly distributed across visual tokens.

    \item \textbf{Focus phase.}
    In the intermediate layers, $R^{(\ell)}$ sharply increases and reaches its peak. We refer to this interval as the focus phase because attention progressively concentrates on a small subset of tokens, thereby most clearly separating strongly supported tokens from weakly supported ones.

    \item \textbf{Rediffusion phase.}
    In the later layers, $R^{(\ell)}$ decreases again, indicating that the previously concentrated attention pattern becomes more distributed.
\end{itemize}

The detailed boundary determination procedure is provided in Appendix~\ref{app:phase_boundary}. This diffusion--focus--rediffusion pattern provides a compact view of how visual token interactions evolve across layers.
The focus phase is especially important because it is where the encoder most clearly separates strongly supported tokens from weakly supported ones. We next examine whether low-attention tokens in this phase can still influence the attention output through their value content.

\subsection{Low-Attention Value Influence in the Focus Phase}
\label{sec:value_amplification}

Attention concentration describes how attention mass is distributed, but it does not directly measure how much value content from a token is reflected in the attention output.
For each layer $\ell$ and head $h$, self-attention computes
\begin{equation}
\mathbf{O}^{(\ell,h)}
=
\mathbf{A}^{(\ell,h)}
\mathbf{V}^{(\ell,h)},
\end{equation}
where $\mathbf{A}^{(\ell,h)}$ is the attention matrix and $\mathbf{V}^{(\ell,h)}$ is the value matrix.
Thus, even tokens receiving small attention mass can still affect the output if their value content remains non-negligible after weighted aggregation.

We define the low-attention token set $\mathcal{L}^{(\ell)}$ as the bottom $25\%$ of visual tokens ranked by received attention mass.
To quantify whether this group still influences the attention output, we measure three quantities:
$\alpha_{\mathcal L}^{(\ell)}$, the total attention mass assigned to low-attention tokens;
$\eta_{\mathcal L}^{(\ell)}$, their normalized value contribution to the attention output;
and
\begin{equation}
G_{\mathcal L}^{(\ell)}
=
\frac{\eta_{\mathcal L}^{(\ell)}}{\alpha_{\mathcal L}^{(\ell)}+\epsilon},
\end{equation}
which measures value influence per unit attention mass.
Detailed definitions of these quantities, including head-wise aggregation, are provided in Appendix~\ref{app:value_influence_metric}.

Fig.~\ref{fig:attn_value_dynamics}(b) shows the layer-wise behavior of these quantities.
In the diffusion phase, $\alpha_{\mathcal L}$ and $\eta_{\mathcal L}$ vary together, yielding a relatively small $G_{\mathcal L}$.
In the focus phase, however, $\alpha_{\mathcal L}$ drops sharply as attention concentrates on a small subset of tokens, while $\eta_{\mathcal L}$ remains non-negligible.
As a result, $G_{\mathcal L}$ increases substantially.
This indicates that low-attention tokens do not dominate the attention output, but their value content exerts a disproportionate influence relative to their small attention mass. Additional low-attention value influence plots for Qwen-2.5-VL and Shikra-7B are provided in Appendix~\ref{app:additional_low_attention_influence}.

This provides a more precise interpretation of the focus phase.
It is not important simply because token values are large.
Rather, it is the stage where attention most clearly separates strongly and weakly supported tokens, while the value content of weakly supported tokens can still be reflected in the attention output.
This attention-value mismatch motivates the phase-wise value intervention analysis in the next section.
\section{Phase-Wise Value Interventions and Hallucination Behavior}
\label{sec:value_intervention}

We now test whether the focus-phase low-attention value signals identified in Sec.~\ref{sec:attention_value_dynamics} are related to object hallucination.
To this end, we conduct controlled value interventions across the diffusion, focus, and rediffusion phases.
We use these interventions as diagnostic tools and measure their effects on caption-level hallucination, token-level object likelihood, and decoder-side visual attention. Unless otherwise specified, the diagnostic experiments in this section are conducted on LLaVA-1.5-7B.

\subsection{Caption-Level Effects of Phase-Wise Value Intervention}
\label{sec:caption_level_value}

\begin{figure}[t]
\centering
\hspace*{0cm}
\includegraphics[width=\textwidth]{value_interventions.pdf}

\caption{
\textbf{Phase-wise value interventions and hallucination behavior.}
We compare four diagnostic interventions across diffusion, focus, and rediffusion phases:
high-attention zero replacement, low-attention zero replacement, low-attention mean replacement, and cross-image value shuffle.
Low-attention value interventions in the focus phase produce the strongest reduction in $\mathrm{CHAIR}_{S}$ and $\mathrm{CHAIR}_{I}$ while largely preserving F1. Metric definitions are provided in Appendix~\ref{Appen:A}.
}
\label{fig:phase_value_intervention}
\vspace{-10pt}
\end{figure}

We partition visual tokens into high- and low-attention groups using the received attention mass defined in Sec.~\ref{sec:value_amplification}.
As shown in Fig.~\ref{fig:phase_value_intervention}, for each phase, we evaluate four diagnostic value interventions:
(A) replacing high-attention values with zero vectors,
(B) replacing low-attention values with zero vectors,
(C) replacing low-attention values with the image-level mean value vector, and
(D) replacing low-attention values with low-attention values from another image.
The high-attention intervention tests the effect of removing strongly supported visual evidence, while the three low-attention interventions remove or disrupt token-specific value content from weakly supported tokens.

We apply each intervention independently to the diffusion, focus, and rediffusion phases, and evaluate
generated captions using CHAIR$_S$, CHAIR$_I$, and F1.
CHAIR$_S$ and CHAIR$_I$ measure sentence-level and instance-level object hallucination, respectively, while F1 measures caption fidelity; detailed benchmark definitions are provided in Appendix~\ref{Appen:A}.
As shown in Fig.~\ref{fig:phase_value_intervention}, interventions in the diffusion and rediffusion phases produce only limited changes in hallucination metrics.
In contrast, modifying low-attention values during the focus phase consistently reduces hallucination.
Low-attention zero replacement, mean replacement, and cross-image value shuffle all reduce $\mathrm{CHAIR}_{S}$ and $\mathrm{CHAIR}_{I}$ while largely preserving F1.

This shared trend is important.
The three low-attention interventions use different replacement values, but they share one common effect: they remove or disrupt the original token-specific value content of focus-phase low-attention tokens.
Thus, the reduction is unlikely to be caused by a particular replacement value.
Rather, it suggests that the original value content of these weakly supported tokens contains hallucination-sensitive signals.



\begin{wraptable}{r}{0.50\textwidth}
\vspace{-15pt}
\centering
\caption{\textbf{Matched controls in the focus phase.}
The phase, 25\% token budget, and mean-replacement operation are fixed, while only the selected token group is varied.}
\label{tab:matched_control_focus}
\vspace{3pt}
\scriptsize
\setlength{\tabcolsep}{3.5pt}
\renewcommand{\arraystretch}{1.08}

\resizebox{\linewidth}{!}{%
\begin{tabular}{lcccc}
\toprule
Metric
& Baseline
& Random
& High
& Ours \\
\midrule

CHAIR$_S \downarrow$
& 45.0
& $39.6 \pm 2.3$
& 42.4
& \textbf{30.8} \\

CHAIR$_I \downarrow$
& 13.3
& $12.6 \pm 0.5$
& 18.1
& \textbf{9.6} \\

\bottomrule
\end{tabular}%
}

\vspace{3pt}
\raggedright
\scriptsize
Ours vs. Random/High: $p<.001$ for both CHAIR$_S$ and CHAIR$_I$.
Full phase-wise results and 95\% CIs are provided in Appendix~\ref{Appen:C.1.1}.
\vspace{-8pt}
\end{wraptable}

The contrast with high-attention intervention further supports this interpretation. In the focus phase, high-attention zero replacement substantially degrades F1, indicating that strongly attended values
contain important grounded visual evidence. By contrast, modifying low-attention values reduces hallucination without a comparable loss in caption fidelity. To further examine whether this effect depends on the selection of low-attention tokens, we perform matched controls that keep the intervention phase, 25\% token budget, and mean-replacement operation fixed while varying only the selected token group. As shown in Table~\ref{tab:matched_control_focus}, low-attention selection in the focus phase yields substantially lower CHAIR$_S$ and CHAIR$_I$ than either random or high-attention selection ($p<0.001$ for both metrics). Full phase-wise results and 95\% confidence intervals are provided in Appendix~\ref{Appen:C.1.1}.

As a complementary diagnostic, we also test the opposite direction by amplifying the contribution of focus-phase low-attention values while keeping the original attention weights and value vectors unchanged. When the amplification factor increases from 1 to 5, CHAIR$_S$ increases from 45.0 to 57.6 and CHAIR$_I$ from 13.3 to 46.5, while F1 decreases from 74.2 to 34.7. The full modulation sweep, statistical trend analysis, and matched amplification controls are provided
in Appendix~\ref{Appen:C.1.2}.

Together, these results indicate that the observed effect is not simply a consequence of applying value intervention to visual tokens or of using a particular replacement rule, but is closely associated with focus-phase low-attention value signals.

\subsection{Token-Level Verification with Teacher Forcing}
\label{sec:teacher_forcing}

Caption-level metrics show that hallucination decreases after focus-phase low-attention value intervention, but they do not directly reveal whether hallucinated object tokens themselves become less likely.
We therefore perform token-level teacher-forcing analysis.

For captions generated by the original model, we identify hallucinated object tokens and ground-truth object tokens.
We then recompute their log-probabilities under the same token sequence while applying focus-phase value interventions.
For an object token span $y_s,\ldots,y_e$, we compute the average log-probability
\begin{equation}
\mathrm{LP}
=
\frac{1}{e-s+1}
\sum_{t=s}^{e}
\log p(y_t \mid \mathbf{x}, y_{<t}),
\end{equation}
where $\mathbf{x}$ is the input image and $y_{<t}$ is the fixed textual prefix.
We report the log-probability change relative to the original model.

\begin{table}[t]
\centering
\caption{
\textbf{Focus-phase low-attention value interventions.}
Hallu LogP $\Delta$ denotes the average log-probability change of hallucinated object tokens relative to the original model.
Drop Rate denotes the percentage of samples where hallucinated-token log-probability decreases.
GT LogP $\Delta$ denotes the corresponding change for ground-truth object tokens. Metric definitions are provided in Appendix~\ref{Appen:A}.
}
\label{tab:phase2_value_intervention}
\small
\setlength{\tabcolsep}{4.5pt}
\renewcommand{\arraystretch}{1.05}
\begin{tabular}{lcccccc}
\toprule
\textbf{Intervention} 
& \textbf{CHAIR$_S$}$\downarrow$ 
& \textbf{CHAIR$_I$}$\downarrow$ 
& \textbf{F1}$\uparrow$ 
& \textbf{Hallu LogP $\Delta$}$\downarrow$ 
& \textbf{Drop Rate}$\uparrow$ 
& \textbf{GT LogP $\Delta$} $\approx 0$ \\
\midrule
\rowcolor{gray!12}
Baseline 
& 45.0 & 13.3 & 74.2 & 0.000 & 0.0 & 0.000 \\
Zero replacement
& 33.4 & 10.6 & 73.6 & $-$0.136 & 63.5 & $-$0.010 \\
Mean replacement
& 30.8 & 9.6 & 74.1 & $-$0.156 & 67.3 & $-$0.020 \\
Cross-image shuffle 
& 31.4 & 10.1 & 73.3 & $-$0.190 & 68.2 & $-$0.052 \\
\bottomrule
\end{tabular}
\end{table}

Table~\ref{tab:phase2_value_intervention} summarizes the results.
All three low-attention value interventions reduce the log-probability of hallucinated object tokens.
Mean replacement decreases hallucinated-token log-probability by $-0.156$ on average, while cross-image value shuffle yields a decrease of $-0.190$.
The drop rate exceeds $67\%$, indicating that hallucinated-token probability decreases in a majority of samples.

The effect on ground-truth object tokens is much smaller.
For mean replacement, the ground-truth log-probability change is only $-0.020$, and F1 remains close to the baseline.
This suggests that the intervention does not globally suppress object generation.
Rather, it preferentially reduces hallucinated object likelihood while largely preserving visually grounded object evidence.

Among the tested interventions, mean replacement provides the best trade-off between hallucination reduction and caption fidelity.
It achieves the lowest $\mathrm{CHAIR}_{S}$ and $\mathrm{CHAIR}_{I}$ while maintaining F1 nearly identical to the baseline.
We therefore use focus-phase low-attention mean replacement as the default intervention in the full experiments.

\subsection{Decoder-Side Visual Attention Analysis}
\label{sec:var_analysis}

\begin{figure}[t]
\centering
\hspace*{0cm}
\includegraphics[width=\textwidth]{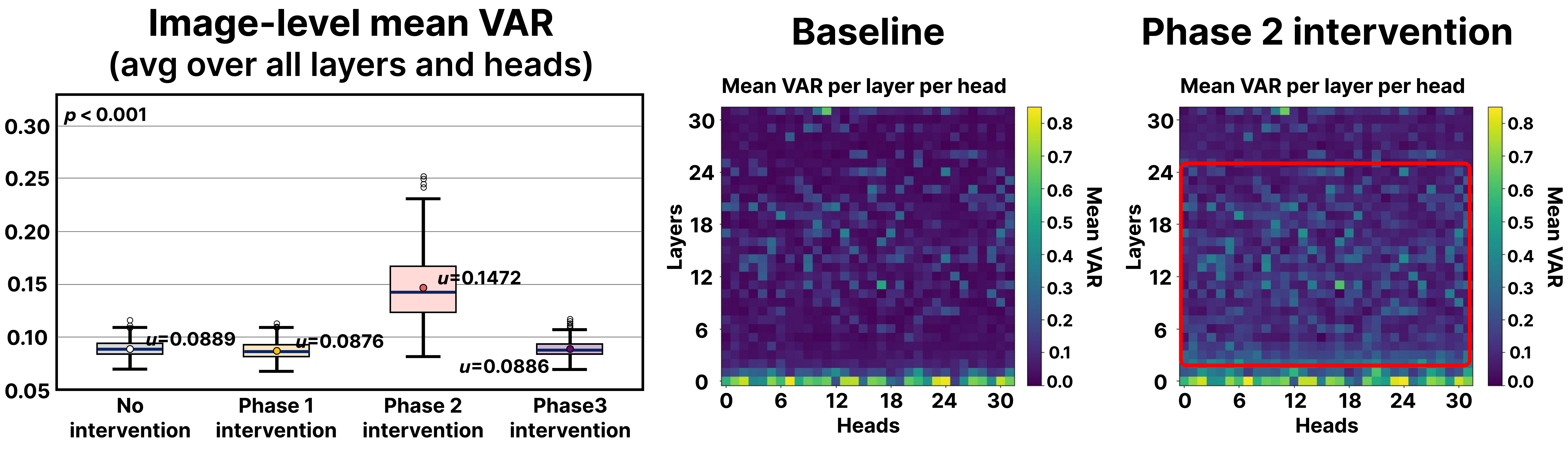}
\caption{
\textbf{Visual Attention Ratio analysis under phase-wise value interventions.}
\textbf{Left:} Distribution of image-level mean VAR across intervention settings.
Focus-phase value intervention yields a significantly higher VAR compared to the baseline ($p<0.001$).
\textbf{Right:} Layer-head VAR heatmaps showing increased visual attention in intermediate language-model layers after focus-phase value intervention.
}
\label{fig:var_analysis}
\vspace{-10pt}
\end{figure}

We further analyze how focus-phase value intervention affects the language model's use of visual tokens during decoding.
Following prior work~\cite{Devils}, we use the Visual Attention Ratio (VAR), which measures the total attention weight assigned by a generated token to image tokens.
For a generated token $y_k$ at language-model layer $l$ and head $h$, VAR is defined as
\begin{equation}
\mathrm{VAR}^{(l,h)}(y_k)
=
\sum_{i=1}^{N}
A^{(l,h)}_k(a_{k,i}),
\end{equation}
where $A^{(l,h)}_k(a_{k,i})$ denotes the attention score from the generated token to the $i$-th visual token.

As shown in Fig.~\ref{fig:var_analysis}, focus-phase value intervention is accompanied by a significant increase in image-level mean VAR, from $0.0889$ to $0.1472$ ($p<0.001$).
In contrast, diffusion- and rediffusion-phase interventions produce negligible changes, with mean VAR values of $0.0876$ and $0.0886$, respectively.
The layer-head heatmaps further show increased attention to visual tokens across multiple language-model layers and heads after focus-phase intervention.

Together with the CHAIR and teacher-forcing results, this suggests that neutralizing focus-phase low-attention values reduces hallucinated object likelihood and is accompanied by stronger decoder-side visual attention, while largely preserving ground-truth object information.

\section{Experiments}
\label{Sec:5}

Based on the attention-value analysis in Sec.~\ref{sec:attention_value_dynamics} and the phase-wise value intervention study in Sec.~\ref{sec:value_intervention}, we evaluate the proposed focus-phase low-attention value replacement strategy. 
In the main paper, we focus on three aspects: hallucination mitigation and compatibility with existing methods, inference efficiency, and broader evaluation and capability preservation.

\subsection{Experimental Setup}
\label{Main:5.1} 

\textbf{Models.}
We evaluate our method on four LVLM backbones with different architectures and scales: LLaVA-1.5-7B, LLaVA-1.5-13B~\cite{llava1_5}, Qwen-2.5-VL-7B~\cite{Qwen2.5-VL}, and InternVL-2.5-8B~\cite{internVL2_5}. 
LLaVA-1.5-7B and LLaVA-1.5-13B are widely used LVLMs based on CLIP~\cite{CLIP} visual features, while Qwen-2.5-VL and InternVL-2.5 represent more recent LVLM architectures. 
Unless otherwise specified, all models are evaluated under standard greedy decoding.

\textbf{Baselines.}
We compare our method with the original model output (Orig.) and Adversarial Uncertainty Estimation (AUE)~\cite{AUE}. 
AUE estimates unreliable visual tokens through iterative adversarial optimization and suppresses their influence during visual encoding. 
We also compare with four decoder-side hallucination mitigation methods—OPERA, VCD, PAI, and Devils—and assess compatibility by reporting each method alone, with AUE, and with ours.

\textbf{Implementation Details.}
We select low-attention visual tokens using the average received attention scores computed from the layers immediately preceding the focus phase. In the focus phase, we replace the selected value vectors with an image-level mean value vector, computed per image, layer, and attention head by averaging over all visual tokens. We use a fixed 25\% ratio across all backbones for the main hallucination benchmarks without model-specific tuning, while the broader MMHal-Bench evaluation uses nearby ratios in the 20--25\% range. Appendix~\ref{Appen:C.2} shows that the effect is generally robust to nearby ratios, with larger ratios trading caption fidelity for stronger CHAIR reduction. All LLaVA-1.5-13B experiments in Table~\ref{tab:combined_comparison} and all MME experiments in Fig.~\ref{fig:mme_radar} were run on an NVIDIA RTX PRO 6000 (96GB); the remaining experiments were run on an NVIDIA RTX 5090 (32GB).

\textbf{Benchmarks and Metrics.}
We evaluate object hallucination using CHAIR~\cite{CHAIR} and POPE~\cite{POPE}.
CHAIR evaluates hallucination in image captioning and reports sentence-level
CHAIR$_S$, instance-level CHAIR$_I$, and F1 for caption fidelity. POPE evaluates object hallucination through binary object-existence questions, and we report F1 scores on the Random, Popular, and Adversarial splits.



For broader hallucination evaluation, we use MMHal-Bench~\cite{MMhal-Bench}, which covers attributes, relations, and reasoning. We report its average score, hallucination rate, and attribute- and relation-level scores. Finally, we use MME~\cite{MME} to assess general multimodal capabilities across perception and cognition tasks. Detailed protocols and metric definitions are provided in Appendix~\ref{Appen:A}.

\begin{table*}[t]
\centering
\caption{
\textbf{Cross-model comparison and compatibility with existing mitigation methods.}
Yellow cells indicate our method. Parentheses report the AUE-relative difference
$\Delta=\text{Ours}-\text{AUE}$, or $+\text{Ours}-(+\text{AUE})$ in (b).
Lower CHAIR$_S$/CHAIR$_I$ and higher F1/POPE F1 are better;
blue deltas favor our method and red deltas favor AUE.
Metric definitions are provided in Appendix~\ref{Appen:A}.
}
\label{tab:combined_comparison}
\small
\setlength{\tabcolsep}{3.2pt}
\renewcommand{\arraystretch}{1.10}

\begin{minipage}{\textwidth}
\vspace{1mm}
\centering
\textbf{(a) Comparison with AUE across LVLMs, including inference time.}\par
\vspace{1mm}
\resizebox{\linewidth}{!}{%
\begin{tabular}{llccc ccc ccc ccc}
\toprule
\multirow{2}{*}{\textbf{Benchmark}}
& \multirow{2}{*}{\textbf{Metric}}
& \multicolumn{3}{c}{\textbf{LLaVA-1.5-7B}}
& \multicolumn{3}{c}{\textbf{LLaVA-1.5-13B}}
& \multicolumn{3}{c}{\textbf{Qwen-2.5-VL}}
& \multicolumn{3}{c}{\textbf{InternVL-2.5}} \\
\cmidrule(lr){3-5}
\cmidrule(lr){6-8}
\cmidrule(lr){9-11}
\cmidrule(lr){12-14}
&
& Orig. & AUE & \cellcolor{oursbg}Ours ($\Delta$)
& Orig. & AUE & \cellcolor{oursbg}Ours ($\Delta$)
& Orig. & AUE & \cellcolor{oursbg}Ours ($\Delta$)
& Orig. & AUE & \cellcolor{oursbg}Ours ($\Delta$) \\
\midrule

\rowcolor{timebg}
& Time (s)$\downarrow$
& 3.207 & 7.307 & \textbf{3.034}
& 6.387 & 14.992 & \textbf{6.366}
& 2.598 & 33.462 & \textbf{2.708}
& 2.299 & 41.291 & \textbf{2.441} \\

\midrule
\multirow{3}{*}{\textbf{CHAIR}}
& CHAIR-S$\downarrow$
& 45.0 & 30.2 & \cellcolor{oursbg}30.8(\textcolor{red}{+0.6})
& 41.0 & 29.8 & \cellcolor{oursbg}32.2(\textcolor{red}{+2.4})
& 31.4 & 27.8 & \cellcolor{oursbg}29.8(\textcolor{red}{+2.0})
& 28.6 & 28.4 & \cellcolor{oursbg}27.8(\textcolor{blue}{-0.6}) \\

& CHAIR-I$\downarrow$
& 13.3 & 10.4 & \cellcolor{oursbg}9.6(\textcolor{blue}{-0.8})
& 11.8 & 9.8 & \cellcolor{oursbg}9.9(\textcolor{red}{+0.1})
& 7.6 & 7.4 & \cellcolor{oursbg}6.9(\textcolor{blue}{-0.5})
& 7.1 & 7.0 & \cellcolor{oursbg}6.9(\textcolor{blue}{-0.1}) \\

& F1$\uparrow$
& 74.2 & 71.8 & \cellcolor{oursbg}74.1(\textcolor{blue}{+2.3})
& 75.3 & 71.4 & \cellcolor{oursbg}73.8(\textcolor{blue}{+2.4})
& 75.5 & 75.8 & \cellcolor{oursbg}75.9(\textcolor{blue}{+0.1})
& 76.3 & 76.6 & \cellcolor{oursbg}76.3(\textcolor{red}{-0.3}) \\

\midrule
\multirow{3}{*}{\textbf{POPE}}
& Random$\uparrow$
& 87.4 & 85.6 & \cellcolor{oursbg}86.4(\textcolor{blue}{+0.8})
& 86.6 & 82.4 & \cellcolor{oursbg}86.01(\textcolor{blue}{+3.61})
& 81.3 & 79.6 & \cellcolor{oursbg}81.4(\textcolor{blue}{+1.8})
& 94.27 & 94.17 & \cellcolor{oursbg}94.4(\textcolor{blue}{+0.23}) \\

& Popular$\uparrow$
& 84.3 & 83.0 & \cellcolor{oursbg}84.0(\textcolor{blue}{+1.0})
& 83.9 & 82.4 & \cellcolor{oursbg}83.47(\textcolor{blue}{+1.07})
& 80.8 & 79.2 & \cellcolor{oursbg}80.8(\textcolor{blue}{+1.6})
& 88.68 & 88.53 & \cellcolor{oursbg}89.3(\textcolor{blue}{+0.77}) \\

& Adversarial$\uparrow$
& 79.3 & 78.4 & \cellcolor{oursbg}79.6(\textcolor{blue}{+1.2})
& 79.5 & 76.5 & \cellcolor{oursbg}78.13(\textcolor{blue}{+1.63})
& 80.5 & 78.7 & \cellcolor{oursbg}80.4(\textcolor{blue}{+1.7})
& 89.6 & 85.73 & \cellcolor{oursbg}85.7(\textcolor{red}{-0.03}) \\
\bottomrule
\end{tabular}%
}
\end{minipage}

\par\vspace{3mm}

\begin{minipage}{\textwidth}
\centering
\textbf{(b) Comparative performance and compatibility with decoder-side hallucination mitigation methods.}\par
\vspace{1mm}
\resizebox{\linewidth}{!}{%
\begin{tabular}{llccc ccc ccc ccc}
\toprule
\multirow{2}{*}{\textbf{Benchmark}}
& \multirow{2}{*}{\textbf{Metric}}
& \multicolumn{3}{c}{\textbf{OPERA}}
& \multicolumn{3}{c}{\textbf{VCD}}
& \multicolumn{3}{c}{\textbf{PAI}}
& \multicolumn{3}{c}{\textbf{Devils}} \\
\cmidrule(lr){3-5}
\cmidrule(lr){6-8}
\cmidrule(lr){9-11}
\cmidrule(lr){12-14}
&
& Orig. & +AUE & \cellcolor{oursbg}+Ours ($\Delta$)
& Orig. & +AUE & \cellcolor{oursbg}+Ours ($\Delta$)
& Orig. & +AUE & \cellcolor{oursbg}+Ours ($\Delta$)
& Orig. & +AUE & \cellcolor{oursbg}+Ours ($\Delta$) \\
\midrule

\multirow{3}{*}{\textbf{CHAIR}}
& CHAIR-S$\downarrow$
& 44.2 & 29.8 & \cellcolor{oursbg}32.6(\textcolor{red}{+2.8})
& 46.2 & 34.2 & \cellcolor{oursbg}39.2(\textcolor{red}{+5.0})
& 32.4 & 19.4 & \cellcolor{oursbg}22.6(\textcolor{red}{+3.2})
& 23.2 & 13.0 & \cellcolor{oursbg}13.4(\textcolor{red}{+0.4}) \\

& CHAIR-I$\downarrow$
& 13.5 & 11.2 & \cellcolor{oursbg}10.6(\textcolor{blue}{-0.6})
& 14.9 & 11.8 & \cellcolor{oursbg}12.9(\textcolor{red}{+1.1})
& 9.62 & 7.61 & \cellcolor{oursbg}7.50(\textcolor{blue}{-0.11})
& 7.88 & 7.24 & \cellcolor{oursbg}5.90(\textcolor{blue}{-1.34}) \\

& F1$\uparrow$
& 74.4 & 72.3 & \cellcolor{oursbg}73.5(\textcolor{blue}{+1.2})
& 71.3 & 70.5 & \cellcolor{oursbg}70.7(\textcolor{blue}{+0.2})
& 74.6 & 67.7 & \cellcolor{oursbg}71.8(\textcolor{blue}{+4.1})
& 72.0 & 63.5 & \cellcolor{oursbg}67.7(\textcolor{blue}{+4.2}) \\

\midrule
\multirow{3}{*}{\textbf{POPE}}
& Random$\uparrow$
& 86.3 & 84.5 & \cellcolor{oursbg}85.5(\textcolor{blue}{+1.0})
& 83.7 & 78.9 & \cellcolor{oursbg}82.1(\textcolor{blue}{+3.2})
& 87.4 & 85.8 & \cellcolor{oursbg}86.1(\textcolor{blue}{+0.3})
& 87.7 & 84.5 & \cellcolor{oursbg}86.0(\textcolor{blue}{+1.5}) \\

& Popular$\uparrow$
& 83.3 & 81.9 & \cellcolor{oursbg}82.8(\textcolor{blue}{+0.9})
& 80.9 & 77.1 & \cellcolor{oursbg}79.5(\textcolor{blue}{+2.4})
& 84.5 & 83.2 & \cellcolor{oursbg}83.9(\textcolor{blue}{+0.7})
& 85.3 & 83.0 & \cellcolor{oursbg}84.3(\textcolor{blue}{+1.3}) \\

& Adversarial$\uparrow$
& 79.3 & 77.5 & \cellcolor{oursbg}78.4(\textcolor{blue}{+0.9})
& 77.3 & 73.3 & \cellcolor{oursbg}76.0(\textcolor{blue}{+2.7})
& 79.7 & 78.8 & \cellcolor{oursbg}79.4(\textcolor{blue}{+0.6})
& 79.9 & 78.7 & \cellcolor{oursbg}79.8(\textcolor{blue}{+1.1}) \\
\bottomrule
\end{tabular}%
}
\vspace{-10pt}
\end{minipage}
\end{table*}

\subsection{Experimental Results}

\textbf{Hallucination mitigation performance and compatibility.} Table~\ref{tab:combined_comparison} summarizes the hallucination mitigation performance of our method across multiple LVLM backbones and its compatibility with existing decoder-side mitigation methods. As shown in Table~\ref{tab:combined_comparison} (a), across all four backbones, our method consistently reduces both CHAIR$_S$ and CHAIR$_I$ relative to the original models while largely preserving caption-level F1. This trend holds for both LLaVA-1.5-7B/13B and more recent architectures such as Qwen-2.5-VL and InternVL-2.5. Compared with AUE, the two methods exhibit different trade-offs: AUE often achieves stronger CHAIR$_S$ reduction, whereas our method generally preserves F1 and POPE performance more effectively and frequently achieves comparable or lower CHAIR$_I$. For example, on LLaVA-1.5-7B, our method reduces CHAIR$_I$ from 13.3 to 9.6 while keeping F1 nearly unchanged (74.2 $\rightarrow$ 74.1), whereas AUE reduces F1 to 71.8.

Table~\ref{tab:combined_comparison} (b) further shows that our vision-encoder intervention can be combined with decoder-side methods including OPERA, VCD, PAI, and Devils. Although the additional CHAIR reduction varies across combinations, +Ours improves hallucination metrics over the corresponding standalone baselines and consistently preserves higher F1 and POPE performance than +AUE across these settings. These results indicate that the proposed intervention can serve as a lightweight complementary module rather than being tied to a particular decoding strategy.

\textbf{Efficiency.} As shown in Table~\ref{tab:combined_comparison} (a), our method introduces only negligible inference overhead compared with the original models. In contrast, AUE relies on iterative adversarial optimization and therefore incurs substantial latency. On Qwen-2.5-VL, AUE increases inference time from 2.598\,s to 33.462\,s, whereas our method requires only 2.708\,s. Similarly, on InternVL-2.5, AUE increases runtime from 2.299\,s to 41.291\,s, while ours requires 2.441\,s. This substantial efficiency advantage stems from the single-pass nature of our intervention, which requires neither iterative optimization nor additional training. Its negligible overhead is particularly beneficial when combined with decoder-side mitigation methods, as it adds little additional latency on top of their existing inference cost.

\begin{table}[t]
\centering
\caption{
\textbf{Broader hallucination evaluation on MMHal-Bench.}
}
\label{tab:broader_hallucination}
\vspace{2pt}

\scriptsize
\setlength{\tabcolsep}{5pt}
\renewcommand{\arraystretch}{0.75}

\resizebox{0.87\textwidth}{!}{%
\begin{tabular}{lcccccccc}
\toprule
\multirow{2}{*}{Model}
& \multicolumn{2}{c}{Avg. Score $\uparrow$}
& \multicolumn{2}{c}{Hallu. Rate $\downarrow$}
& \multicolumn{2}{c}{Attribute $\uparrow$}
& \multicolumn{2}{c}{Relation $\uparrow$} \\
\cmidrule(lr){2-3}
\cmidrule(lr){4-5}
\cmidrule(lr){6-7}
\cmidrule(lr){8-9}
& Orig. & Ours
& Orig. & Ours
& Orig. & Ours
& Orig. & Ours \\
\midrule

LLaVA-1.5-13B
& 2.04 & \textbf{2.08}
& 0.604 & \textbf{0.604}
& 2.42 & \textbf{3.08}
& 1.58 & \textbf{1.83} \\

Qwen-2.5-VL
& 4.18 & \textbf{4.21}
& 0.260 & \textbf{0.240}
& 3.83 & \textbf{3.83}
& 4.50 & \textbf{4.58} \\

InternVL-2.5
& 3.73 & \textbf{3.83}
& 0.260 & \textbf{0.229}
& 3.42 & \textbf{3.67}
& 4.17 & \textbf{4.17} \\

\bottomrule
\end{tabular}%
}

\vspace{-7pt}
\end{table}









\textbf{Broader evaluation and capability preservation.} To examine whether the effect extends beyond object-existence hallucination, we further evaluate our method on MMHal-Bench using LLaVA-1.5-13B, Qwen-2.5-VL-7B, and InternVL-2.5-8B. For this broader evaluation, we use low-attention token ratios ranging from 20\% to 25\%, while keeping the remaining intervention settings unchanged. As shown in Table~\ref{tab:broader_hallucination}, the average score improves across all three backbones, while all reported attribute- and relation-level scores are maintained or improved. The hallucination rate also decreases on Qwen-2.5-VL and InternVL-2.5 while remaining unchanged on LLaVA-1.5-13B. Together, these results indicate that the effect of our intervention extends beyond object-existence errors to broader attribute- and relation-related hallucination settings.

\begin{wrapfigure}{r}{0.48\linewidth}
\vspace{-8pt}
\centering
\includegraphics[width=\linewidth]{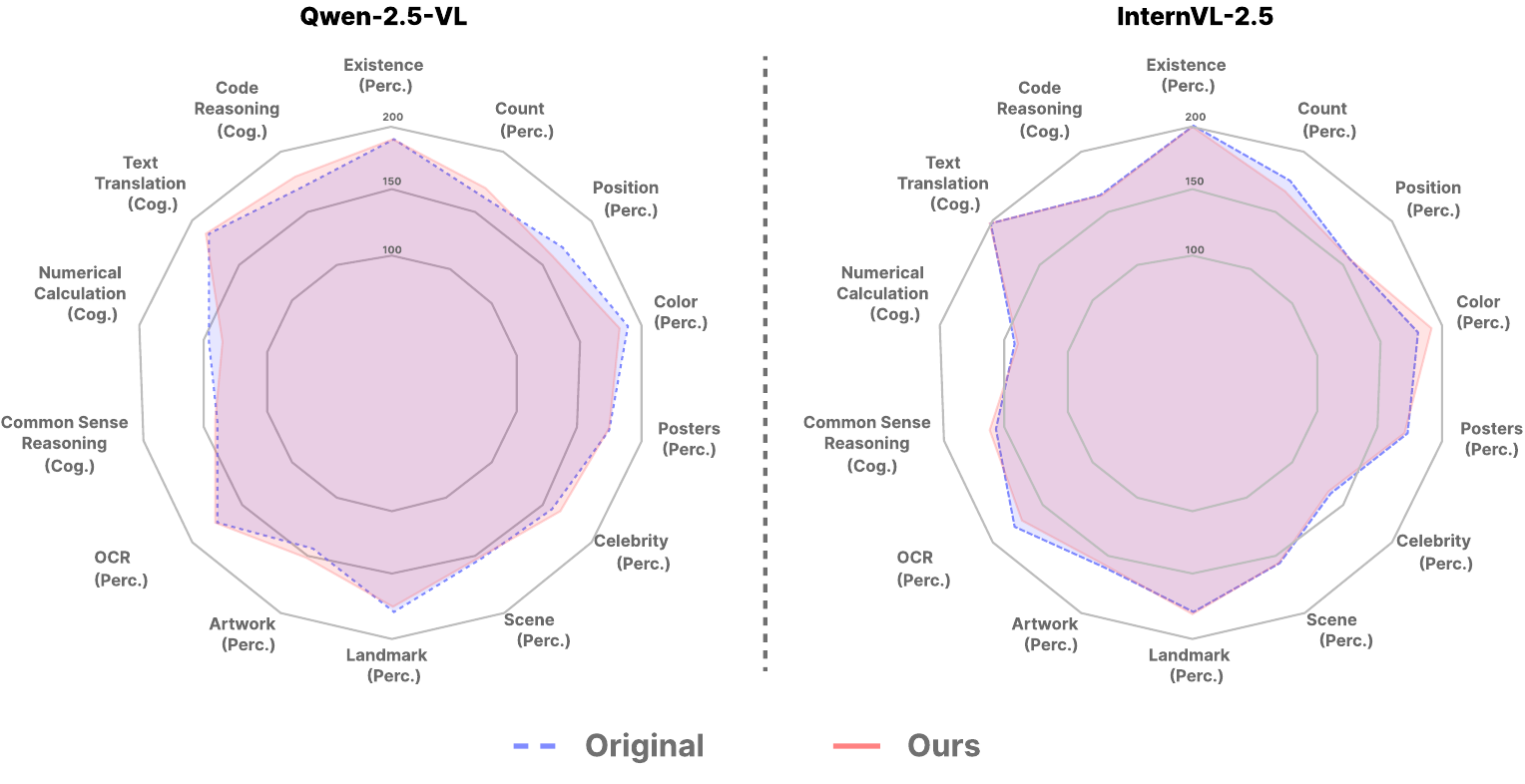}
\vspace{-6pt}
\caption{
\textbf{Category-level MME comparison between the original models and our method.}
Our method largely preserves the original MME performance profile across Qwen-2.5-VL and InternVL-2.5.
}
\label{fig:mme_radar}
\vspace{-12pt}
\end{wrapfigure}
We additionally evaluate general multimodal capability on MME. As shown in Fig.~\ref{fig:mme_radar}, our method largely preserves the original category-level performance profiles of Qwen-2.5-VL and InternVL-2.5 across both perception- and cognition-oriented categories. Although small category-wise fluctuations are observed, the overall profiles remain similar, suggesting that the intervention mitigates hallucination without broadly degrading general multimodal capability.

Overall, focus-phase low-attention value replacement provides a favorable balance of hallucination reduction, output fidelity, and inference efficiency across diverse LVLMs, while remaining compatible with decoder-side mitigation and largely preserving 
general multimodal capabilities.

\section{Conclusion}

In this paper, we studied object hallucination in LVLMs through the lens of attention-value dynamics inside vision encoders. Our analysis identified a consistent diffusion--focus--rediffusion structure and showed that the focus phase is the stage where attention most clearly separates strongly supported tokens from weakly supported ones. We further found that low-attention tokens in this phase can still exert non-negligible value-side influence relative to their small attention mass, revealing an attention-value mismatch that is not captured by attention scores alone.
Through phase-wise value interventions, we showed that modifying focus-phase low-attention values reduces hallucination metrics while largely preserving caption fidelity.
Token-level teacher-forcing analysis further demonstrated that this intervention decreases hallucinated object likelihood with much smaller effects on ground-truth object tokens.
Motivated by these observations, we used a simple training-free intervention that replaces focus-phase low-attention values with an image-level mean value vector.
Across multiple LVLM backbones, this intervention reduces object hallucination with negligible additional runtime.
These findings suggest that hallucination mitigation can benefit from analyzing not only which visual tokens receive attention, but also how value-side signals from weakly supported tokens are propagated through the vision encoder.
\newpage
\begin{ack}
This work was supported by the Institute of Information \& Communications Technology Planning \& Evaluation (IITP) grant funded by the Korea government (MSIT) (IITP-2026-RS-2026-25615271, AI Star Fellowship Support Program). Computational resources were provided by “Advanced GPU Utilization Support Program” funded by the Government of the Republic of Korea (Ministry of Science and ICT) and the Cluster Server for Computational Science at Pusan National University.
\end{ack} 



\bibliographystyle{ieeetr}
\bibliography{ref}







\newpage
\section*{NeurIPS Paper Checklist}

\begin{enumerate}

\item {\bf Claims}
    \item[] Question: Do the main claims made in the abstract and introduction accurately reflect the paper's contributions and scope?
    \item[] Answer: \answerYes{} 
    \item[] Justification: Our approach to hallucination mitigation is sufficiently described in the abstract and introduction. In particular, we summarize our contributions at the end of the introduction.

\item {\bf Limitations}
    \item[] Question: Does the paper discuss the limitations of the work performed by the authors?
    \item[] Answer: \answerYes{} 
    \item[] Justification: We discuss the limitations of our work in Appendix~\ref{Appen:D}

\item {\bf Theory assumptions and proofs}
    \item[] Question: For each theoretical result, does the paper provide the full set of assumptions and a complete (and correct) proof?
    \item[] Answer: \answerNA{} 
    \item[] Justification: The paper does not present formal theoretical results, theorems, or proofs. 
The equations in the paper are used to define analysis metrics and intervention procedures, and their assumptions and definitions are described in the main text and supplementary material.

    \item {\bf Experimental result reproducibility}
    \item[] Question: Does the paper fully disclose all the information needed to reproduce the main experimental results of the paper to the extent that it affects the main claims and/or conclusions of the paper (regardless of whether the code and data are provided or not)?
    \item[] Answer: \answerYes{} 
    \item[] Justification: We provide code for reproducibility in the supplementary material, and we describe the methodology, benchmarks, and baseline models used in our experiments in detail in Sec~\ref{Main:5.1}.

\item {\bf Open access to data and code}
    \item[] Question: Does the paper provide open access to the data and code, with sufficient instructions to faithfully reproduce the main experimental results, as described in supplemental material?
    \item[] Answer: \answerYes{} 
    \item[] Justification: We provide our code and reproduction instructions in the supplementary material. All evaluation data are from publicly available benchmarks, including COCO val2014, POPE, MME, and MMHal-Bench, and we describe the data access, preprocessing, and evaluation settings needed to reproduce the main results.

\item {\bf Experimental setting/details}
    \item[] Question: Does the paper specify all the training and test details (e.g., data splits, hyperparameters, how they were chosen, type of optimizer) necessary to understand the results?
    \item[] Answer: \answerYes{} 
    \item[] Justification: We specify the evaluated models, benchmarks, data splits, decoding settings, intervention hyperparameters, and runtime measurement details in Sec.~\ref{Main:5.1}. 
Since our method is training-free, no optimizer or training hyperparameters are required.

\item {\bf Experiment statistical significance}
    \item[] Question: Does the paper report error bars suitably and correctly defined or other appropriate information about the statistical significance of the experiments?
    \item[] Answer: \answerYes{} 
    \item[] Justification: We report statistical significance for the VAR analysis and provide 95\% confidence intervals and two-sided p-values using paired image-level bootstrap resampling for matched mean-replacement controls \ref{Appen:C.1.1}. We further report exact two-sided permutation Spearman tests for continuous value modulation and Holm-corrected significance results for matched amplification controls \ref{Appen:C.1.2}. Other benchmark evaluations use frozen models, deterministic decoding, fixed random seeds, and fixed evaluation splits; therefore, we do not report run-to-run error bars for every benchmark table.

\item {\bf Experiments compute resources}
    \item[] Question: For each experiment, does the paper provide sufficient information on the computer resources (type of compute workers, memory, time of execution) needed to reproduce the experiments?
    \item[] Answer: \answerYes{} 
    \item[] Justification: We specify the GPU specifications used for each experiment in main paper \ref{Main:5.1}, where one RTX 5090 and one RTX PRO 6000 were used.
    
\item {\bf Code of ethics}
    \item[] Question: Does the research conducted in the paper conform, in every respect, with the NeurIPS Code of Ethics \url{https://neurips.cc/public/EthicsGuidelines}?
    \item[] Answer: \answerYes{} 
    \item[] Justification: We comply with the NeurIPS Code of Ethics.

\item {\bf Broader impacts}
    \item[] Question: Does the paper discuss both potential positive societal impacts and negative societal impacts of the work performed?
    \item[] Answer: \answerYes{} 
    \item[] Justification: We discuss that reducing object hallucination can improve the reliability of LVLMs, while also noting risks such as over-reliance on imperfect mitigation and potential misuse in sensitive applications. 
These broader impacts are discussed in Appendix~\ref{Appen:E}, and limitations are discussed in Appendix~\ref{Appen:D}.
    
\item {\bf Safeguards}
    \item[] Question: Does the paper describe safeguards that have been put in place for responsible release of data or models that have a high risk for misuse (e.g., pre-trained language models, image generators, or scraped datasets)?
    \item[] Answer: \answerNA{} 
    \item[] Justification: Our paper does not introduce any new models or datasets.

\item {\bf Licenses for existing assets}
    \item[] Question: Are the creators or original owners of assets (e.g., code, data, models), used in the paper, properly credited and are the license and terms of use explicitly mentioned and properly respected?
    \item[] Answer: \answerYes{} 
    \item[] Justification: We cite the original papers for the existing LVLMs, baselines, and benchmark datasets used in our experiments. 
We use these assets for research evaluation only and do not redistribute modified versions of them.

\item {\bf New assets}
    \item[] Question: Are new assets introduced in the paper well documented and is the documentation provided alongside the assets?
    \item[] Answer: \answerYes{} 
    \item[] Justification: We provide a README file in our code, included in the supplementary material, that describes the usage instructions, benchmarks, and models.

\item {\bf Crowdsourcing and research with human subjects}
    \item[] Question: For crowdsourcing experiments and research with human subjects, does the paper include the full text of instructions given to participants and screenshots, if applicable, as well as details about compensation (if any)? 
    \item[] Answer: \answerNA{} 
    \item[] Justification: Our paper does not involve crowdsourcing or research with human subjects.

\item {\bf Institutional review board (IRB) approvals or equivalent for research with human subjects}
    \item[] Question: Does the paper describe potential risks incurred by study participants, whether such risks were disclosed to the subjects, and whether Institutional Review Board (IRB) approvals (or an equivalent approval/review based on the requirements of your country or institution) were obtained?
    \item[] Answer: \answerNA{} 
    \item[] Justification: Our paper does not involve research with human subjects.

\item {\bf Declaration of LLM usage}
    \item[] Question: Does the paper describe the usage of LLMs if it is an important, original, or non-standard component of the core methods in this research? Note that if the LLM is used only for writing, editing, or formatting purposes and does \emph{not} impact the core methodology, scientific rigor, or originality of the research, declaration is not required.
    \item[] Answer: \answerNA{} 
    \item[] Justification: The development of our method does not involve LLMs as any important, original, or non-standard components.

\end{enumerate}

\newpage{}
\appendix
 
\section*{Organization of the Appendices}
\addcontentsline{toc}{section}{Organization of the Appendix}

These appendices provide detailed benchmark definitions, extended attention-value analyses, additional experimental validation, and discussions of limitations and broader impacts to complement the main manuscript.

The document is organized as follows:

\begin{itemize}

\item \textbf{Appendix \ref{Appen:A}} describes detailed benchmark definitions used in our experiments.

\item \textbf{Appendix \ref{Appen:B}} provides details of the attention-value analysis, including phase boundary determination, the definition of the low-attention value influence metric, and additional influence plots on representative backbones.

\item \textbf{Appendix \ref{Appen:C}} presents additional analyses of focus-phase low-attention value intervention, including matched controls and statistical validation, continuous value modulation, sensitivity to the low-attention token ratio, and robustness to focus-phase boundaries.

\item \textbf{Appendix \ref{Appen:D}} discusses the limitations and future directions of our work.

\item \textbf{Appendix \ref{Appen:E}} discusses the broader impacts of our work.

\end{itemize}

\section{Benchmarks}
\label{Appen:A}

\paragraph{CHAIR.}
CHAIR~\cite{CHAIR} evaluates object hallucination in image captioning by comparing generated object mentions with ground-truth object annotations. 
Following the standard protocol, we generate captions for 500 images sampled from COCO val2014 using the prompt ``Please describe this image in detail.'' 
We report sentence-level CHAIR$_S$, instance-level CHAIR$_I$, and F1. 
CHAIR$_S$ measures the proportion of generated captions that contain at least one hallucinated object, while CHAIR$_I$ measures the proportion of hallucinated objects among all mentioned objects:
\begin{equation}
\mathrm{CHAIR}_S =
\frac{|\{\text{captions containing hallucinated objects}\}|}
{|\{\text{all captions}\}|},
\quad
\mathrm{CHAIR}_I =
\frac{|\{\text{hallucinated objects}\}|}
{|\{\text{all mentioned objects}\}|}.
\end{equation}
Lower CHAIR$_S$ and CHAIR$_I$ indicate fewer hallucinations. 
We also report F1 to measure caption fidelity with respect to ground-truth object annotations.

\paragraph{POPE.}
POPE~\cite{POPE} evaluates object hallucination through binary object-existence questions. 
We follow the standard POPE protocol and report F1 scores on the Random, Popular, and Adversarial splits. 
The Random split samples negative objects randomly, the Popular split uses frequently occurring objects as negatives, and the Adversarial split selects more challenging negative objects that are likely to be confused with the image content. 
Higher F1 scores indicate stronger object-level grounding and better resistance to object hallucination.

\paragraph{MMHal-Bench.}
MMHal-Bench~\cite{MMhal-Bench} evaluates hallucination in practical multimodal
question-answering settings across diverse question categories and object topics. Following the standard evaluation protocol, open-ended model responses are assessed for both overall response quality and hallucination. We report the average response score, evaluated on a 0--6 scale, and the hallucination rate, which measures the frequency of hallucinated responses. To further examine broader hallucination behavior, we additionally report scores for attribute- and relation-related categories. Higher average, attribute, and relation scores and a lower hallucination rate indicate better performance.

\paragraph{MME.}
MME~\cite{MME} evaluates general multimodal capabilities beyond hallucination-specific captioning benchmarks. 
We use MME to examine whether the proposed intervention preserves category-level performance across both perception-oriented and cognition-oriented tasks. 
The perception-oriented categories include object existence, count, position, color, OCR, posters, celebrity, scene, artwork, and landmark. 
The cognition-oriented categories include commonsense reasoning, numerical calculation, text translation, and code reasoning. 
We compare the category-level MME profiles of the original models and our method to verify that hallucination mitigation does not globally degrade general multimodal capability.


\section{Details of Attention-Value Analysis}
\label{Appen:B}
 \subsection{Phase Boundary Determination}
\label{app:phase_boundary}
In Sec.~\ref{sec:phase_structure} of the main manuscript, we empirically observed a consistent three-phase visual processing structure---Diffusion, Focus, and Rediffusion---across multiple LVLM vision encoders. 
To clarify how the phase boundaries used in our experiments are determined, we describe the procedure for identifying the focus-phase interval $\mathcal{F}=[\ell_{\mathrm{start}}, \ell_{\mathrm{end}}]$ based on the layer-wise dynamics of the attention concentration ratio.

The concentration ratio at layer $\ell$ is defined as
\begin{equation}
R^{(\ell)} = \frac{M^{(\ell)}}{H^{(\ell)}},
\end{equation}
where $M^{(\ell)}$ denotes the maximum attention score and $H^{(\ell)}$ denotes the attention entropy at layer $\ell$. 
This ratio increases when attention becomes concentrated on a small subset of tokens and decreases when attention is more evenly distributed.

\subsubsection{Onset of the Focus Phase}

During the initial diffusion phase, attention is broadly distributed and the layer-wise change of the concentration ratio remains relatively small. 
To identify the transition into the focus phase, we examine the discrete forward difference
\begin{equation}
\Delta R^{(\ell)} = R^{(\ell)} - R^{(\ell-1)} .
\end{equation}

The start layer $\ell_{\mathrm{start}}$ is determined as the first layer where the increase in concentration exceeds the baseline variation observed in the early layers:
\begin{equation}
\ell_{\mathrm{start}}
=
\min
\left\{
\ell
\mid
\Delta R^{(\ell)}
>
\mu_{\mathrm{base}}
+
\lambda \cdot \sigma_{\mathrm{base}}
\right\},
\end{equation}
where $\mu_{\mathrm{base}}$ and $\sigma_{\mathrm{base}}$ denote the mean and standard deviation of $\Delta R^{(\ell)}$ computed from the early diffusion layers, e.g., the first $25\%$ of the network depth, and $\lambda$ is a confidence multiplier. 
We set $\lambda=2$ for all evaluated models;
using $\lambda=3$ yields identical focus-phase intervals. This criterion identifies the layer where attention concentration begins to increase beyond the relatively stable baseline behavior of the diffusion stage.

\subsubsection{Offset of the Focus Phase}

After identifying $\ell_{\mathrm{start}}$, we determine the end layer of the focus phase by using a proportional window over the vision encoder depth. 
Specifically, the focus-phase length is defined as
\begin{equation}
K = \mathrm{round}(0.3L),
\end{equation}
where $L$ denotes the total number of vision-encoder layers. 
The end layer is then given by
\begin{equation}
\ell_{\mathrm{end}} = \ell_{\mathrm{start}} + K - 1 .
\end{equation}

This proportional window captures the main high-concentration region following the onset of the focus phase, while avoiding model-specific tuning based on downstream hallucination metrics. 
Layers before $\ell_{\mathrm{start}}$ are treated as the diffusion phase, and layers after $\ell_{\mathrm{end}}$ are treated as the rediffusion phase.

\subsubsection{Phase Boundaries Across Evaluated Models}
\begin{figure}[t]
\centering
\hspace*{0cm}
\includegraphics[width=0.98\textwidth]{att_dynamics_supple.pdf}
\caption{\textbf{Layer-wise attention dynamics across evaluated LVLM backbones.} The solid lines indicate the ratio of maximum attention score to entropy ($R^{(l)}$). The shaded regions represent the identified Diffusion (yellow), Focus (red), and Rediffusion (purple) phases based on our gradient-based boundary formulation.
}
\label{fig:Phase}
\vspace{-5pt}
\end{figure}
Applying the above procedure yields the layer intervals used in our experiments, as visualized in Fig.~\ref{fig:Phase}. 
The boundaries are determined from attention concentration dynamics, not from validation performance or hallucination metrics.

\paragraph{LLaVA-1.5-7B and LLaVA-1.5-13B ($L=24$).}
For both LLaVA-1.5 backbones, the concentration change $\Delta R^{(\ell)}$ remains close to zero in early layers and exhibits a clear increase around layer 11. 
With $\ell_{\mathrm{start}}=11$ and $K=\mathrm{round}(0.3L)=7$, the focus-phase interval becomes layers 11--17.

\paragraph{Qwen-2.5-VL ($L=32$).}
For Qwen-2.5-VL, the concentration change exceeds the early baseline around layer 17. 
With $\ell_{\mathrm{start}}=17$ and $K=\mathrm{round}(0.3L)=10$, the focus-phase interval becomes layers 17--26, covering the dominant high-concentration region before the concentration ratio decreases. The sharp dips at indices 7, 15, 23, and 31 coincide with the only layers that use full attention; the remaining layers use window attention. Full attention operates over a broader set of tokens and can temporarily reduce the measured concentration. As seen around indices 22–24, the concentration recovers in neighboring window-attention layers. We therefore interpret these dips as local effects of the attention-pattern switch rather than additional phase boundaries.

\paragraph{InternVL-2.5 ($L=24$).}
For InternVL-2.5, the first consistent increase in $\Delta R^{(\ell)}$ occurs around layer 9. 
With $\ell_{\mathrm{start}}=9$ and $K=\mathrm{round}(0.3L)=7$, the focus-phase interval becomes layers 9--15. 
Although a secondary spike appears in later layers, it is largely confined to CLS-to-patch attention and diminishes when averaged over patch queries. Because InternVL-2.5 discards the CLS token before projecting patch features to the language model, we use the first dominant concentration region, which marks the initial transition from broadly distributed attention to the focus phase.

\subsection{Definition of Low-Attention Value Influence}
\label{app:value_influence_metric}

We provide the detailed formulation of the low-attention value influence metric used in Sec.~\ref{sec:value_amplification}.
For layer $\ell$ and head $h$, let $\mathbf{A}^{(\ell,h)} \in \mathbb{R}^{N \times N}$ denote the visual self-attention matrix and $\mathbf{V}^{(\ell,h)} \in \mathbb{R}^{N \times d_h}$ denote the corresponding value matrix.
The received attention mass of token $j$ is defined as
\begin{equation}
s_j^{(\ell)}
=
\frac{1}{|\mathcal{H}|N}
\sum_{h \in \mathcal{H}}
\sum_{i=1}^{N}
A_{i,j}^{(\ell,h)} .
\end{equation}
We define the low-attention token set $\mathcal{L}^{(\ell)}$ as the bottom $25\%$ of tokens according to $s_j^{(\ell)}$.

The attention mass assigned to the low-attention group is
\begin{equation}
\alpha_{\mathcal{L}}^{(\ell)}
=
\frac{1}{|\mathcal{H}|N}
\sum_{h \in \mathcal{H}}
\sum_{i=1}^{N}
\sum_{j \in \mathcal{L}^{(\ell)}}
A_{i,j}^{(\ell,h)} .
\end{equation}
The attention-output contribution from the low-attention group is
\begin{equation}
\mathbf{O}_{\mathcal{L}}^{(\ell,h)}
=
\mathbf{A}_{:,\mathcal{L}}^{(\ell,h)}
\mathbf{V}_{\mathcal{L}}^{(\ell,h)} .
\end{equation}
We normalize this contribution by the full attention output:
\begin{equation}
\eta_{\mathcal{L}}^{(\ell)}
=
\frac{
\sum_{h \in \mathcal{H}}
\left\|
\mathbf{O}_{\mathcal{L}}^{(\ell,h)}
\right\|_{F}
}{
\sum_{h \in \mathcal{H}}
\left\|
\mathbf{A}^{(\ell,h)}
\mathbf{V}^{(\ell,h)}
\right\|_{F}
+\epsilon
}.
\end{equation}
Finally, the influence per attention mass is defined as
\begin{equation}
G_{\mathcal{L}}^{(\ell)}
=
\frac{
\eta_{\mathcal{L}}^{(\ell)}
}{
\alpha_{\mathcal{L}}^{(\ell)}+\epsilon
}.
\end{equation}
 

\begin{figure}[t]
\centering
\includegraphics[width=\textwidth]{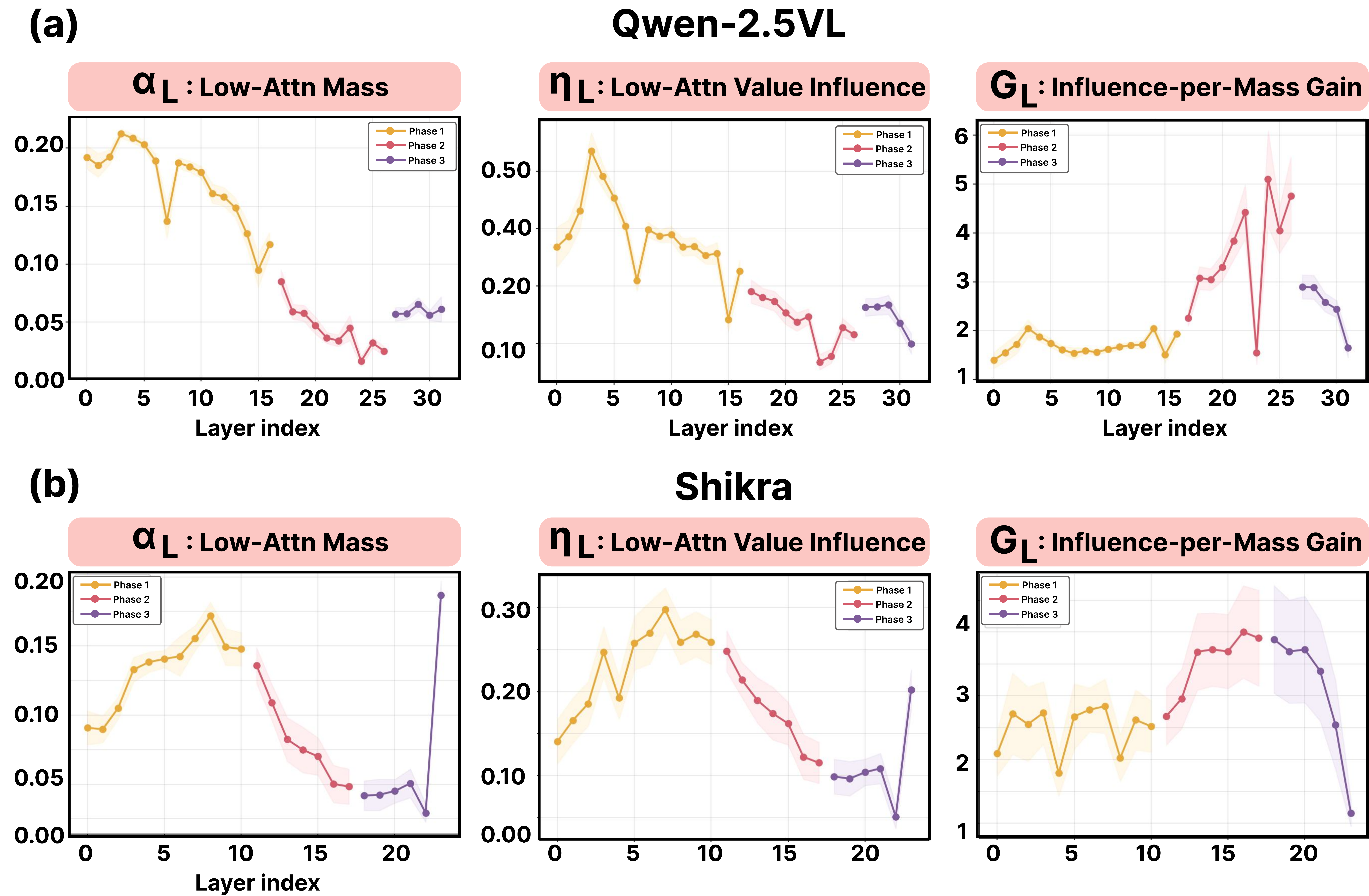}
\caption{
\textbf{Additional low-attention value influence analysis.}
We report $\alpha_{\mathcal L}$, $\eta_{\mathcal L}$, and $G_{\mathcal L}$ for Qwen-2.5-VL and Shikra-7B.
Both models show a similar qualitative trend to the main LLaVA analysis: around the focus phase, the attention mass assigned to low-attention tokens decreases, while their normalized value contribution remains non-negligible, increasing the influence per attention mass.
The strength of the effect varies across architectures.
}
\label{fig:additional_low_attention_influence}
\end{figure}

\subsection{Additional Low-Attention Value Influence Analysis}
\label{app:additional_low_attention_influence}

As shown in Fig.~\ref{fig:additional_low_attention_influence}, we provide additional low-attention value influence plots for Qwen-2.5-VL and Shikra-7B.
Both models show the same qualitative behavior as the main LLaVA analysis: the attention mass assigned to low-attention tokens decreases around the focus phase, while their normalized value contribution remains non-negligible, leading to an increased influence per attention mass.
The magnitude of this effect varies across architectures, and we therefore use these plots as diagnostic evidence rather than as a strict universality claim.

\section{Additional Analysis and Robustness of Focus-Phase Low-Attention Intervention}
\label{Appen:C}

\subsection{Specificity and Statistical Validation of Low-Attention Value Intervention}
\label{Appen:C.1}

\subsubsection{Mean-replacement matched controls}
\label{Appen:C.1.1}

\begin{table}[t]
\centering
\caption{
\textbf{Mean-replacement matched controls across intervention phases.}
We fix the intervention phase, token budget (25\%), and mean-replacement operation,
and vary only the selected visual-token group.
Random results are reported as mean $\pm$ standard deviation.
$\Delta$ denotes Low minus the corresponding control.
}
\label{tab:matched_mean_replacement}
\small
\setlength{\tabcolsep}{4pt}
\renewcommand{\arraystretch}{1.10}

\resizebox{\textwidth}{!}{%
\begin{tabular}{llcccccc}
\toprule
Phase & Metric
& Baseline
& Low
& Random
& High
& Low--Random $\Delta$ [95\% CI], $p$
& Low--High $\Delta$ [95\% CI], $p$ \\
\midrule

\multirow{2}{*}{P1}
& CHAIR$_S \downarrow$
& 45.0
& 44.0
& $45.8 \pm 0.3$
& 44.0
& $-1.8$ [$-5.4$, $+1.9$], $p=.32$
& $+0.0$ [$-4.6$, $+4.4$], $p=1.00$ \\

& CHAIR$_I \downarrow$
& 13.3
& 16.2
& $15.2 \pm 0.3$
& 14.6
& $+1.0$ [$-0.6$, $+2.6$], $p=.22$
& $+1.7$ [$-0.3$, $+3.6$], $p=.10$ \\

\midrule

\multirow{2}{*}{P2}
& CHAIR$_S \downarrow$
& 45.0
& \textbf{30.8}
& $39.6 \pm 2.3$
& 42.4
& $\mathbf{-8.8}$ [$\mathbf{-12.0}$, $\mathbf{-5.6}$], $p<.001$
& $\mathbf{-11.6}$ [$\mathbf{-16.6}$, $\mathbf{-6.6}$], $p<.001$ \\

& CHAIR$_I \downarrow$
& 13.3
& \textbf{9.6}
& $12.6 \pm 0.5$
& 18.1
& $\mathbf{-2.9}$ [$\mathbf{-4.1}$, $\mathbf{-1.8}$], $p<.001$
& $\mathbf{-8.5}$ [$\mathbf{-10.7}$, $\mathbf{-6.3}$], $p<.001$ \\

\midrule

\multirow{2}{*}{P3}
& CHAIR$_S \downarrow$
& 45.0
& 45.2
& $43.9 \pm 1.2$
& 49.4
& $+1.3$ [$-1.5$, $+4.1$], $p=.36$
& $-4.2$ [$-8.4$, $+0.0$], $p=.06$ \\

& CHAIR$_I \downarrow$
& 13.3
& 13.2
& $13.2 \pm 0.3$
& 14.7
& $+0.0$ [$-0.9$, $+1.0$], $p=.95$
& $-1.5$ [$-3.0$, $+0.1$], $p=.06$ \\

\bottomrule
\end{tabular}%
}
\end{table}

In Sec.~\ref{sec:caption_level_value} of the main paper, we showed that hallucination is consistently reduced when the token-specific content of focus-phase low-attention values is removed or disrupted using different intervention strategies. This suggests that the observed effect does not depend on a particular replacement value. Here, we further examine whether the selection of low-attention tokens itself is important for this effect.

To this end, we perform matched control experiments in which the intervention phase, token budget, and replacement operation are fixed, while only the selected visual-token group is varied. Specifically, within each phase, we apply the same mean-replacement operation to the same 25\% token budget and compare three token-selection strategies: \textit{Low}, which selects the bottom 25\% of visual tokens according to received attention mass; \textit{Random}, which uniformly samples 25\% of the visual tokens; and \textit{High}, which selects the top 25\% according to received attention mass. We perform this comparison independently in the diffusion (P1), focus (P2), and rediffusion (P3) phases. For the Random control, we report the mean and standard deviation across repeated runs. For the difference between Low and each matched control, we compute 95\% confidence intervals and two-sided $p$-values using paired image-level bootstrap resampling.

As shown in Table~\ref{tab:matched_mean_replacement}, a significant advantage of low-attention selection emerges only in the focus phase (P2). For CHAIR$_S$, Low achieves 30.8, which is 8.8 points lower than Random at $39.6 \pm 2.3$ (95\% CI: $[-12.0,-5.6]$) and 11.6 points lower than High at 42.4 (95\% CI: $[-16.6,-6.6]$); both differences are significant with $p<0.001$. For CHAIR$_I$, Low achieves 9.6, which is 2.9 and 8.5 points lower than Random at $12.6 \pm 0.5$ and High at 18.1, respectively. Both differences are again statistically significant ($p<0.001$).

In contrast, in the diffusion (P1) and rediffusion (P3) phases, Low does not show a statistically significant advantage over either the Random or High control. These matched controls therefore indicate that the hallucination reduction observed in the main paper is not a generic consequence of applying mean replacement to a fixed number of visual tokens, but is closely associated with selecting low-attention tokens specifically during the focus phase.

\subsubsection{Continuous Modulation of Low-Attention Value Signals}
\label{Appen:C.1.2}


\begin{table}[t]
\centering
\caption{
\textbf{Continuous modulation of low-attention value contribution.}
We amplify the contribution of focus-phase low-attention values by a scalar $\alpha$
while keeping the original value vectors and attention weights unchanged.
As $\alpha$ increases, CHAIR$_S$ and CHAIR$_I$ increase, while F1 decreases.
}
\label{tab:value_modulation}
\footnotesize
\setlength{\tabcolsep}{3.5pt}
\renewcommand{\arraystretch}{1.08}

\begin{tabular}{lccccccc}
\toprule
Metric
& $\alpha=1.0$
& $\alpha=1.5$
& $\alpha=2.0$
& $\alpha=2.5$
& $\alpha=3.0$
& $\alpha=4.0$
& $\alpha=5.0$ \\
\midrule
CHAIR$_S \downarrow$
& 45.0 & 47.4 & 48.4 & 49.0 & 48.0 & 54.2 & 57.6 \\
CHAIR$_I \downarrow$
& 13.3 & 14.2 & 14.7 & 15.8 & 18.4 & 29.0 & 46.5 \\
F1 $\uparrow$
& 74.2 & 73.3 & 72.8 & 70.8 & 67.9 & 56.3 & 34.7 \\
\bottomrule
\end{tabular}
\end{table}

\begin{table}[t]
\centering
\caption{
\textbf{Matched amplification controls.}
Differences in CHAIR$_S$ between low-attention amplification and norm-matched
high-attention or random-token amplification. Positive values indicate a larger
hallucination increase for the low-attention group.
}
\label{tab:matched_amplification}
\small
\setlength{\tabcolsep}{5pt}
\renewcommand{\arraystretch}{1.10}
\begin{tabular}{ccc}
\toprule
$\alpha$
& Low--High $\Delta$ CHAIR$_S$ [95\% CI]
& Low--Random $\Delta$ CHAIR$_S$ [95\% CI] \\
\midrule
1.25 & $+1.8$ [$-1.6$, $+5.2$]  & $+0.9$ [$-1.7$, $+3.4$] \\
1.50 & $\mathbf{+6.6}$ [$+2.8$, $+10.6$] & $+2.9$ [$+0.3$, $+5.9$] \\
1.75 & $\mathbf{+6.4}$ [$+2.0$, $+11.2$] & $\mathbf{+5.2}$ [$+2.3$, $+8.3$] \\
2.00 & $\mathbf{+12.6}$ [$+8.2$, $+17.2$] & $\mathbf{+5.7}$ [$+2.6$, $+8.9$] \\
\bottomrule
\end{tabular}
\end{table}

In the main manuscript, we show that neutralizing focus-phase low-attention values reduces hallucination while largely preserving caption fidelity. 
Here, we perform a complementary diagnostic experiment: instead of suppressing or replacing these values, we progressively amplify their contribution to the attention output and examine how hallucination metrics change.

We conduct this experiment on LLaVA-1.5-7B using 500 images from COCO~\cite{COCO} val2014. 
We identify the low-attention token group using the same criterion as in the main experiments: the bottom $25\%$ visual tokens according to pre-focus received attention mass, with the CLS token protected. 
The modulation is then applied to the focus-phase target layers.

Unlike replacement-based interventions, this experiment keeps the original value vectors unchanged. 
We only modulate the contribution of the low-attention group to the attention output by a scalar $\alpha$:
\begin{equation}
\mathbf{O}(\alpha)
=
\mathbf{A}_{\mathcal{H}}\mathbf{V}_{\mathcal{H}}
+
\alpha \cdot
\mathbf{A}_{\mathcal{L}}\mathbf{V}_{\mathcal{L}},
\end{equation}
where $\mathcal{H}$ denotes the remaining visual tokens and $\mathcal{L}$ denotes the low-attention token group. 
The attention weights and all value vectors are kept unchanged; only the gain applied to the low-attention value contribution is varied. 
We sweep $\alpha \in \{1, 1.5, 2, 2.5, 3, 4, 5\}$, where $\alpha=1$ corresponds to the original model.


As shown in Table~\ref{tab:value_modulation}, increasing the contribution of low-attention values generally worsens hallucination while reducing caption fidelity. As $\alpha$ increases from 1 to 5, CHAIR$_S$ rises from 45.0 to 57.6 and CHAIR$_I$ increases more sharply from 13.3 to 46.5, while F1 decreases from 74.2 to 34.7. Across the seven amplification settings, exact two-sided permutation Spearman tests confirm that these trends are statistically significant: CHAIR$_S$ shows $\rho=0.893$ ($p=0.012$), CHAIR$_I$ shows $\rho=1.000$ ($p<0.001$), and F1 shows $\rho=-1.000$ ($p<0.001$). These results demonstrate a consistent relationship between increasing low-attention value contribution and hallucination, providing complementary evidence to the value-replacement experiments in the main paper.

We next perform matched amplification controls to examine whether this sensitivity is specific to the low-attention group. We amplify the contributions of high-attention and random-token groups while norm-matching the perturbation magnitude to that of the low-attention condition. As shown in Table~\ref{tab:matched_amplification}, low-attention amplification induces a larger increase in CHAIR$_S$ than high-attention and random-token amplification as $\alpha$ increases. After Holm correction, the low-attention condition produces a significantly larger increase in hallucination than the high-attention condition for $\alpha \geq 1.5$, and than the random-token condition for $\alpha \geq 1.75$.

These results indicate that the observed sensitivity is not merely a generic effect of amplification, but is closely associated with focus-phase low-attention value signals. Importantly, this experiment does not modify the token content itself; it only changes the gain applied to their contribution to the attention output. Thus, the results support the interpretation that focus-phase low-attention value signals can influence hallucination behavior without making the stronger claim that all low-attention tokens are inherently harmful.

\subsection{Sensitivity to the Low-Attention Token Ratio}
\label{Appen:C.2}
\begin{table}[t]
\centering
\caption{
\textbf{Sensitivity to the low-attention token ratio.}
We vary the fraction of low-attention tokens used for focus-phase value replacement and report CHAIR$_S$, CHAIR$_I$, and F1.
Larger ratios can further reduce CHAIR metrics, but may introduce a stronger fidelity trade-off depending on the model.
We use $25\%$ as the default ratio because it provides a conservative balance between hallucination reduction and caption fidelity without model-specific tuning.
}
\label{tab:ratio_robustness}
\small
\setlength{\tabcolsep}{4.5pt}
\renewcommand{\arraystretch}{1.10}
\resizebox{\linewidth}{!}{%
\begin{tabular}{lccccccccc}
\toprule
\multirow{2}{*}{Ratio}
& \multicolumn{3}{c}{LLaVA-1.5-7B}
& \multicolumn{3}{c}{Qwen-2.5-VL}
& \multicolumn{3}{c}{InternVL-2.5} \\
\cmidrule(lr){2-4}
\cmidrule(lr){5-7}
\cmidrule(lr){8-10}
& CHAIR$_S$ $\downarrow$ & CHAIR$_I$ $\downarrow$ & F1 $\uparrow$
& CHAIR$_S$ $\downarrow$ & CHAIR$_I$ $\downarrow$ & F1 $\uparrow$
& CHAIR$_S$ $\downarrow$ & CHAIR$_I$ $\downarrow$ & F1 $\uparrow$ \\
\midrule
\rowcolor{gray!10}
Baseline
& 45.0 & 13.3 & 74.2
& 31.4 & 7.6 & 75.5
& 28.6 & 7.1 & 76.3 \\
\midrule
$20\%$
& 36.2 & 11.3 & 74.0
& 30.0 & 7.4 & 75.2
& 29.0 & 7.5 & 76.3 \\

$25\%$
& 30.8 & 9.6 & 74.1
& 29.8 & 6.9 & 75.9
& 27.8 & 6.9 & 76.3 \\

$30\%$
& 27.8 & 9.4 & 72.5
& 28.4 & 6.7 & 76.1
& 27.2 & 6.6 & 76.3 \\

$35\%$
& 25.4 & 8.8 & 70.4
& 26.8 & 6.7 & 76.0
& 26.0 & 6.8 & 76.2 \\

$40\%$
& 22.6 & 8.0 & 69.1
& 27.8 & 6.8 & 76.4
& 27.2 & 7.3 & 75.8 \\
\bottomrule
\end{tabular}%
}
\end{table}

In the main experiments, we use the bottom $25\%$ visual tokens as the low-attention group for focus-phase value replacement. 
To examine how the intervention behaves under different token ratios, we vary the replacement ratio from $20\%$ to $40\%$ and evaluate CHAIR$_S$, CHAIR$_I$, and F1 across multiple LVLM backbones.

As shown in Table~\ref{tab:ratio_robustness}, increasing the ratio generally strengthens hallucination suppression, but the fidelity trade-off differs across models. 
On LLaVA-1.5-7B, larger ratios further reduce hallucination, with CHAIR$_S$ decreasing from $30.8$ at $25\%$ to $22.6$ at $40\%$, and CHAIR$_I$ decreasing from $9.6$ to $8.0$. 
However, this stronger hallucination reduction comes with a clear F1 drop from $74.1$ to $69.1$, indicating that aggressive replacement may neutralize value signals that remain useful for caption generation.

For Qwen-2.5-VL, CHAIR metrics remain lower than or close to the original baseline across all ratios, while F1 changes only marginally.
On InternVL-2.5, ratios between $25\%$ and $35\%$ consistently reduce CHAIR metrics while maintaining F1 close to the baseline; at $40\%$, the CHAIR$_I$ improvement becomes weaker.
These results suggest that the intervention is not dependent on a finely tuned ratio, although overly aggressive replacement can introduce a stronger fidelity trade-off.

Based on these observations, we use $25\%$ as the default ratio in the main experiments. 
Although larger ratios can sometimes produce stronger CHAIR reduction, especially for LLaVA-1.5-7B and Qwen-2.5-VL, they may also increase the risk of fidelity degradation depending on the model. 
The $25\%$ setting therefore provides a conservative balance between hallucination reduction and caption fidelity without model-specific ratio tuning.

\subsection{Robustness to Focus-Phase Boundaries}
\label{Appen:C.3}

\begin{table}[t]
\centering
\caption{
\textbf{Robustness to focus-phase boundaries.}
We perturb the default focus-phase interval by contracting or expanding both boundaries
by one layer while keeping all other intervention settings fixed.
}
\label{tab:focus_boundary_robustness}
\small
\setlength{\tabcolsep}{5pt}
\renewcommand{\arraystretch}{1.08}

\begin{tabular}{llcccc}
\toprule
Model & Setting & Target layers
& CHAIR$_S \downarrow$
& CHAIR$_I \downarrow$
& F1 $\uparrow$ \\
\midrule

\multirow{4}{*}{LLaVA-1.5-7B}
& \cellcolor{gray!15} No intervention
& \cellcolor{gray!15} --
& \cellcolor{gray!15} 45.0
& \cellcolor{gray!15} 13.3
& \cellcolor{gray!15} 74.2 \\
& Contracted & [12--16] & 33.0 & 10.3 & 73.6 \\
& \cellcolor{yellow!15} Default
& \cellcolor{yellow!15} [11--17]
& \cellcolor{yellow!15} 30.8
& \cellcolor{yellow!15} 9.6
& \cellcolor{yellow!15} 74.1 \\
& Expanded & [10--18] & 32.0 & 10.0 & 72.3 \\
\midrule

\multirow{4}{*}{Qwen-2.5-VL-7B}
& \cellcolor{gray!15} No intervention
& \cellcolor{gray!15} --
& \cellcolor{gray!15} 31.4
& \cellcolor{gray!15} 7.6
& \cellcolor{gray!15} 75.5 \\
& Contracted & [18--25] & 28.2 & 7.1 & 76.1 \\
& \cellcolor{yellow!15} Default
& \cellcolor{yellow!15} [17--26]
& \cellcolor{yellow!15} 29.8
& \cellcolor{yellow!15} 6.9
& \cellcolor{yellow!15} 75.9 \\
& Expanded & [16--27] & 26.0 & 6.6 & 76.7 \\
\midrule

\multirow{4}{*}{InternVL-2.5-8B}
& \cellcolor{gray!15} No intervention
& \cellcolor{gray!15} --
& \cellcolor{gray!15} 28.6
& \cellcolor{gray!15} 7.1
& \cellcolor{gray!15} 76.3 \\
& Contracted & [10--14] & 27.4 & 6.9 & 76.3 \\
& \cellcolor{yellow!15} Default
& \cellcolor{yellow!15} [9--15]
& \cellcolor{yellow!15} 27.8
& \cellcolor{yellow!15} 6.9
& \cellcolor{yellow!15} 76.3 \\
& Expanded & [8--16] & 27.8 & 6.9 & 75.8 \\
\bottomrule
\end{tabular}
\end{table}

In the main paper, we define the focus phase of each model based on its attention dynamics and apply the low-attention value intervention within the corresponding layer range. Here, we examine whether the observed hallucination reduction is sensitive to the exact focus-phase boundaries by shifting the start and end boundaries of the default focus interval inward or outward by one layer. All other token-selection and intervention settings are kept identical to those used in the main experiments.

As shown in Table~\ref{tab:focus_boundary_robustness}, the effect of the intervention is largely preserved under these boundary perturbations. For LLaVA-1.5-7B, both CHAIR$_S$ and CHAIR$_I$ remain lower than the no-intervention baseline not only for the default interval $[11,17]$, but also for the contracted $[12,16]$ and expanded $[10,18]$ settings. The same trend is observed for Qwen-2.5-VL-7B and InternVL-2.5-8B, where both CHAIR metrics remain below the corresponding baselines across all tested boundary settings.

Performance under the contracted and expanded intervals also remains broadly comparable to that of the default setting, while F1 is generally maintained at a similar level to the original models. These results indicate that the effectiveness of the proposed intervention does not depend strongly on a single exact layer boundary and remains stable under small perturbations around the identified focus phase.

\section{Limitations and Future Directions}
\label{Appen:D}

Despite its favorable hallucination reduction--efficiency trade-off, our method has several limitations.
First, focus-phase low-attention value replacement can weaken visual signals that are useful for fine-grained perception.
As shown in the ratio analysis, increasing the replacement ratio can further reduce CHAIR metrics but may also decrease F1, indicating a trade-off between hallucination suppression and caption fidelity.
Therefore, a fixed replacement ratio may not be optimal for all images, prompts, or downstream tasks.
Developing adaptive image- or prompt-dependent replacement strategies is an important direction for future work.

Second, our analysis provides diagnostic evidence rather than a formal causal proof of hallucination mechanisms.
Although the diffusion--focus--rediffusion pattern is observed across multiple LVLM backbones, the phase boundaries are determined by an empirical attention-concentration criterion rather than a theoretical definition.
A more principled characterization of phase transitions inside vision encoders could further clarify why the focus phase is particularly sensitive to low-attention value-side signals.

Third, our evaluations mainly focus on image-based object hallucination.
In more complex settings such as video understanding, 3D perception, multi-image reasoning, and dense spatial reasoning, low-attention tokens may contain small, rare, or contextually important visual evidence.
In such cases, more careful adaptive interventions may be required to reduce hallucination while preserving task-relevant information.

Finally, improvements in generative hallucination metrics do not always directly translate to improvements in binary object-existence evaluation.
For example, POPE changes can be mixed across splits, suggesting that reducing hallucinated object mentions and improving yes/no object grounding are related but not identical objectives.
Future work could study how value-side interventions affect calibration and object-existence decision boundaries.

\section{Broader impacts}
\label{Appen:E}

Our work aims to improve the reliability of LVLMs by reducing object hallucination in image-grounded generation. 
This can have positive impacts in applications that require faithful visual grounding, such as assistive technologies, educational tools, robotics, and human-AI decision support systems. 
By reducing unsupported object mentions, hallucination mitigation can make LVLM outputs easier to interpret and less likely to mislead users.

However, our method does not eliminate hallucinations completely and should not be viewed as a guarantee of factual correctness. 
Users may over-rely on model outputs if hallucination mitigation is mistaken for complete reliability, especially in sensitive domains such as medical, legal, robotics, or safety-critical decision-making. 
In addition, deployment of LVLMs in surveillance or automated monitoring can raise privacy and fairness concerns that are outside the scope of our method. 
Thus, the proposed intervention should be used with appropriate human oversight, task-specific validation, and safeguards in high-stakes settings.

\end{document}